\documentclass[sigconf,screen,noacm]{acmart}

\usepackage{xspace}
\usepackage{amsmath}

\usepackage{amssymb}
\usepackage{hyperref}
\usepackage{tabularx}
\usepackage{tikz}
\usetikzlibrary{shapes.geometric, arrows.meta, positioning, calc, fit}
\usepackage{listings}
\usepackage{enumitem}
\setlist[itemize]{leftmargin=1.2em}
\setlist[enumerate]{leftmargin=1.2em}

\newcommand{\TriCEGAR}{{\text{TriCEGAR}}\xspace} 

\newif\ifnotes
\notestrue 

\newcommand{\Eta}{\mathrm{H}}

\AtBeginDocument{%
  }

\setcopyright{none}
\copyrightyear{}
\acmYear{}
\acmDOI{}
\acmISBN{}
\acmConference{}
\acmBooktitle{}
\acmPrice{}

\begin{document}

\title{\TriCEGAR: A Trace-Driven Abstraction Mechanism for Agentic AI}

\author{Roham Koohestani}
\authornote{Equal contribution}
\affiliation{
  \institution{JetBrains Research}
  \city{Amsterdam}
  \country{Netherlands}
}
\email{roham.koohestani@jetbrains.com}

\author{Ateş Görpelioğlu}
\authornotemark[1]
\affiliation{
  \institution{Delft University of Technology}
  \city{Delft}
  \country{Netherlands}
}
\email{agorpelioglu@tudelft.nl}

\author{Egor Klimov}
\affiliation{
  \institution{JetBrains Research}
  \city{Amsterdam}
  \country{Netherlands}
}
\email{egor.klimov@jetbrains.com}

\author{Burcu Kulahcioglu Ozkan}
\affiliation{
  \institution{Delft University of Technology}
  \city{Delft}
  \country{Netherlands}
}
\email{b.ozkan@tudelft.nl}

\author{Maliheh Izadi}
\affiliation{
  \institution{Delft University of Technology}
  \city{Delft}
  \country{Netherlands}
}
\email{m.izadi@tudelft.nl}

\renewcommand{\shortauthors}{Koohestani et al.}

\begin{abstract}
Agentic AI systems act through tools and evolve their behavior over long,
stochastic interaction traces. This setting complicates assurance, because
behavior depends on nondeterministic environments and probabilistic
model outputs. Prior work introduced runtime verification for agentic AI
via \emph{Dynamic Probabilistic Assurance} (DPA), learning an MDP online and
model checking quantitative properties. A key limitation is that developers
must manually define the state abstraction, which couples verification to
application-specific heuristics and increases adoption friction.
This paper proposes \TriCEGAR, a trace-driven abstraction mechanism that
automates state construction from execution logs and supports online
construction of an agent behavioral MDP. \TriCEGAR represents abstractions
as predicate trees learned from traces and refined using counterexamples.
We describe a framework-native implementation that (i) captures typed agent
lifecycle events, (ii) builds abstractions from traces,
(iii) constructs an MDP, and (iv) performs probabilistic model checking to
compute bounds such as $P_{\max}(\Diamond \mathsf{success})$ and
$P_{\min}(\Diamond \mathsf{failure})$. We also show how run likelihoods
enable anomaly detection as a guardrailing signal.
\end{abstract}



\maketitle

\section{Introduction}
\label{sec:introduction}

There has been a recent shift in attention from large language models (LLMs) toward autonomous agentic systems~\cite{xi_rise_2023}. Whereas LLMs have largely served as reactive oracles, agentic systems are built to sense evolving environments, reason over multi-step objectives, and perform actions via external tool interfaces~\cite{mcp}. This expanded capability enables applications spanning automated software repair~\cite{bouzenia_repairagent_2024} to sophisticated financial decision-making~\cite{xiao_tradingagents_2025}. Yet this autonomy creates new challenges for providing assurances about such systems: the probabilistic nature of their neural foundations, together with their ability to act without strict constraints, makes conventional deterministic verification techniques insufficient.

Runtime verification provides an alternative by checking properties against
executions and reacting during operation.
Recent work~\cite{koohestani_agentguard_2025} introduced \emph{Dynamic Probabilistic Assurance} (DPA) for
agentic AI, learning an MDP online from interaction traces and verifying
quantitative properties using probabilistic model checking.
However, the state abstraction in existing prototypes requires manual
rules mapping raw logs to formal states.
This manual step is labor-intensive, brittle under agent changes, and
limits reuse across agents. Additionally, it functions as a middleware layer without integration into enterprise-grade agent frameworks like Koog~\cite{jetbrainskoog_2025}, CrewAI~\cite{crewai}, or LangGraph~\cite{langgraph}, which deliver structural guarantees through typed event loops and persistent state management. 

This paper targets the next step: \emph{automating} state abstraction for
runtime verification of agentic AI.
We propose \TriCEGAR, a dynamic counter-example-guided approach for semantic state abstraction,
which enables the building of an Agentic MDP (AMDP).
We additionally report an implementation that integrates trace capture
into an agent framework lifecycle and supports end-to-end generation of
probability bounds and anomaly signals.

The primary contributions of our work include: (1) \TriCEGAR, a trace-driven abstraction mechanism that replaces manual
    state mapping with learned predicate trees and counterexample-guided
    refinement. (2) A framework-native implementation that captures the agent lifecycle
    events, builds abstractions via decision trees, constructs an MDP, and
    model checks PCTL properties to compute quantitative assurances.
    (3) A guardrailing signal based on run likelihood under the induced
    probabilistic model for anomaly detection.
    (4) An actionable roadmap for future extensions of the idea and framework.
An early version of the implementation is made available through the replication package\footnote{\href{https://zenodo.org/records/18338703}{https://zenodo.org/records/18338703}} and supports end-to-end model learning and chekcing.


\section{Motivation}
\label{sec:motivation}

Assurance for agentic systems differs fundamentally from assurance for deterministic software components due to the stochastic nature of the underlying models. While deterministic systems may fail due to logic errors, agentic systems act probabilistically, meaning they can exhibit transient failures even when the underlying logic is sound. Specifically, the nondeterministic generation of tool calls and arguments increases the likelihood that an agent will (i) select tools incorrectly based on ambiguous context, (ii) enter probabilistic loops where steps are repeated without progress, (iii) violate workflow constraints (e.g., committing without tests) due to alignment failures, or (iv) exhibit non-stationary behavior when prompts, tools, or environments change.
These failures occur during execution and often require runtime intervention.

Two requirements follow.
First, developers need \emph{high-level, behavioral} properties that are
meaningful for agent workflows, rather than low-level assertions on code.
Second, the assurance mechanism must adapt when the agent evolves, which
requires the state representation used for verification to be derived from
traces rather than manually maintained rules.
\TriCEGAR targets these requirements by constructing and refining a formal
state space from observed executions, enabling verification queries that
return quantitative bounds and can trigger warnings when risk increases.


\section{Background}
\label{sec:background}

\textbf{Runtime Verification.}
Runtime verification (RV)~\cite{bartocci_lectures_2018, DBLP:series/lncs/BartocciFFR18} analyzes executions against a specification, often
by attaching monitors to event streams. Monitor-oriented programming separates
business logic from monitoring logic, which supports integrating monitors
alongside an agent execution loop~\cite{chen_mop_2007,meredith_overview_2012}. In agentic settings, the event stream
includes tool invocations, observations, intermediate artifacts, and internal memory updates.
RV is typically viewed as complementary to offline verification~\cite{bayazit2005complementary,leucker2009brief}. The monitor
operates under partial information and must decide what to observe and how to
summarize it~\cite{cimatti2019assumption}. For agentic systems, the summarization problem is a state
abstraction problem~\cite{li2006towards}, mapping a high-dimensional execution context
into a tractable state space on which properties can be evaluated.
More specifically, in our case, we are dealing with a model-irrelevance abstraction
based on the taxonomy in~\cite{li2006towards}.

\textbf{Probabilistic Model Checking.}
Probabilistic model checking (PMC)~\cite{baier2008principles} extends verification to models with
randomized transitions, which make quantitative statements about likelihoods.
In an MDP, nondeterminism captures choices (e.g., environment
uncertainty), and probability captures stochastic
outcomes. Properties are commonly expressed using PCTL, including reachability
queries such as:
$
  P_{\max}=?\ [\ \Diamond\ \mathsf{success}\ ],
$
and
$
  P_{\min}=?\ [\ \Diamond\ \mathsf{failure}\ ].
$
Bounds are useful in runtime, as they expose uncertainty and can
be used as risk indicators when environments are partially modeled.

\textbf{Dynamic Probabilistic Assurance.}
Dynamic Probabilistic Assurance (DPA)~\cite{koohestani_agentguard_2025} builds an MDP online from
observations and applying PMC to compute quantitative assurances during
execution. AgentGuard implements this
idea by observing agent I/O, abstracting it into events, learning an MDP of behavior, and model checking safety or success properties in real
time.
A limiting factor is the reliance on manually defined abstractions, which
map raw execution context into states. Manual abstractions are costly
to develop, must be updated when agents evolve, and are sensitive to changes
in tool interfaces and memory schemas. \TriCEGAR addresses this gap by learning
and refining abstractions from traces.

\textbf{CEGAR and Probabilistic CEGAR.}
CEGAR~\cite{cegar} iteratively refines an abstraction using counterexamples that are
spurious under the concrete system. Extending CEGAR to probabilistic systems
requires defining counterexamples for quantitative properties and using
diagnostic information to refine the abstraction~\cite{hermanns2008probabilistic}. 
Existing approaches refine abstractions to tighten
probability bounds or eliminate spurious witnesses, often using interval-based
abstractions, stochastic games, or critical subsystems
~\cite{de2007magnifying, kattenbelt2010game,han2009counterexample}.
\TriCEGAR adopts this principle, but couples refinement with a trace-driven
predicate tree that is learned from observed agent executions.


\section{\TriCEGAR}

\subsection{Approach}
\label{sec:approach}

\begin{figure*}
    \centering
    \includegraphics[width=0.90\linewidth]{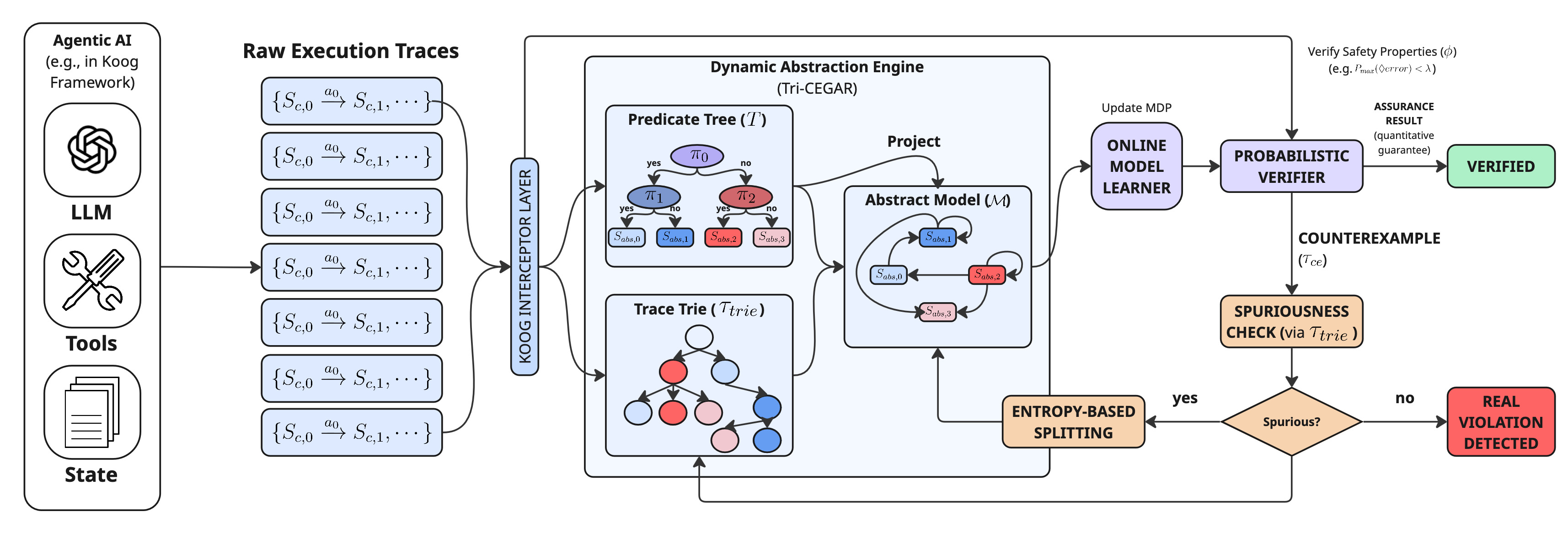}
    \caption{Visualization of the \TriCEGAR approach for online AMDP learning.}
    \label{fig:tricegar_pipeline}
\end{figure*}

This subsection details \TriCEGAR, a framework for reducing the manual effort required to verify agentic systems. As illustrated in ~\autoref{fig:tricegar_pipeline}, the approach operates in a cycle: it captures raw execution traces, learns a predicate tree to abstract these traces into states, constructs an MDP from the abstract transitions, and verifies properties against this model. If verification yields a counterexample that is spurious (i.e., not supported by the concrete traces), the system refines the abstraction tree to separate distinct behaviors.

\subsubsection{Setting and Objectives}
Let $\Sigma$ denote the set of concrete agent states captured from runtime snapshots
(memory variables, control flags, tool inputs, and observations). A trace is
a sequence
$
  \sigma = (\sigma_0 \xrightarrow{a_0} \sigma_1 \xrightarrow{a_1} \cdots \xrightarrow{a_{n-1}} \sigma_n),
$
where $\sigma_i \in \Sigma$ is a concrete state and $a_i$ is a tool invocation (action).

Given a set of traces $L$ and a property $\phi$ (for example, success
reachability or failure avoidance), \TriCEGAR aims to:
\begin{enumerate}
  \item construct an abstraction $\alpha:\Sigma \rightarrow S$ which maps a concrete state 
  to an abstract state in the finite abstract state space $S$,
  \item construct an MDP $\mathcal{M}$ over $S$ from abstracted traces, and
  \item compute quantitative guarantees via PMC, then refine $\alpha$ when
  evidence indicates that $\mathcal{M}$ is too coarse to be stable
\end{enumerate}

The induced model is represented as an MDP whose actions correspond to tool
types 
(e.g., \texttt{git\_commit},
\texttt{run\_test}). In this representation, nondeterminism captures uncertainty 
about the deployed agent's tool-selection policy, while probabilities capture
stochastic outcomes of executing a selected tool from an abstract state.
As a result, quantities such as $P_{\max}$ and $P_{\min}$ provide upper and
lower envelopes over admissible tool-selection strategies, and can be used
as conservative risk indicators. Separately, for anomaly detection we score
an observed run using the realized action sequence via the action-conditioned
likelihood in \autoref{sec:implementation:anomaly}.
\TriCEGAR supports online operation: traces arrive incrementally during execution, 
the abstraction and MDP are updated between steps, and the monitor 
continually refines the probability bounds.

\subsubsection{Three Linked Structures}
\TriCEGAR maintains three linked structures, which together define
the current verification state:
\begin{enumerate}
  \item \textbf{Predicate tree $T$} defines an abstraction function $\alpha_T$
  by routing each concrete state $s_c$ from the root to a leaf based on
  predicate outcomes. Leaves correspond to abstract states.
  \item \textbf{Trace trie $\mathcal{T}$} stores observed abstract trace
  prefixes. This supports (i) efficient lookups of realizable prefixes and
  (ii) mapping abstract witnesses back to observed behaviors.
  \item \textbf{Constructed model $\mathcal{M}$} is an MDP whose states correspond
  to leaves of $T$, actions correspond to tools, and transition
  probabilities are estimated from trace frequencies.
\end{enumerate}

The key design choice is to keep these structures consistent: each leaf in
$T$ corresponds to an MDP vertex in $\mathcal{M}$ and is associated with a
set of trie endpoints that reference the concrete states mapped to that
leaf. This enables refinement actions that are local to a leaf, but
also update both the constructed model and trace indexing.

\subsubsection{Initialization via Decision-Tree Learning}
Given an initial log $L_{\mathrm{init}}$, \TriCEGAR learns $T_0$ using a greedy
decision tree procedure. Each candidate predicate $\pi$ partitions a multiset
of concrete states $B$ into $B_{\mathsf{true}}$ and $B_{\mathsf{false}}$.
Predicates are selected by information gain (IG) with respect to the next-action
distribution:
\[
  \mathrm{IG}(B,\pi)=
  \mathcal{H}(A \mid B) -
  \sum_{b \in \{\mathsf{true},\mathsf{false}\}}
  \frac{|B_b|}{|B|} \mathcal{H}(A \mid B_b),
\]
where $\mathcal{H}$ is Shannon entropy over the empirical next-action
distribution. The next-action proxy is chosen as tool choice is a
directly observable behavioral signal in tool-using agents and is available
without requiring outcome labels.
This initialization yields a non-trivial state space that already separates
distinct action regimes, which reduces refinements compared to a
single-state start.

\subsubsection{Predicate Grammar}
\TriCEGAR draws predicates from a typed grammar that is grounded in captured
variables and their types:
\vspace{-0.3em}
\begin{itemize}
  \item \textbf{Scalar predicates} for numeric variables, using thresholds
  derived from observed values (for example, quantiles or midpoints between
  sorted distinct values): $\mathsf{iterations}>\theta$.
  \item \textbf{Boolean predicates} for control-flow flags:
  $\mathsf{testsPassed}=\mathsf{true}$.
  \item \textbf{Structural predicates} for collections and artifacts:
  $\mathsf{diffs}\neq\emptyset$.
  \item \textbf{Semantic predicates} for text variables, when embeddings are
  available. Given an embedding $E(v)\in\mathbb{R}^d$, a predicate is
  $\mathbf{w}^\top E(v)+b \ge 0$. We currently do not
  support this type of predicates by they are in active development (Section~\ref{sec:implementation}).
\end{itemize}
\vspace{-0.3em}

\subsubsection{Verification and Refinement Loop}
At runtime, the monitor periodically: (1) projects new concrete states through $\alpha_T$ to update counts and
  extend $\mathcal{T}$, (2) updates transition estimates in $\mathcal{M}$, and (3) checks property $\phi$ on $\mathcal{M}$ to get updated probability bounds.
When model checking returns a violation and provides a diagnostic witness,
\TriCEGAR interprets this result using the trace trie. If the
witness includes an abstract transition that has no supporting evidence
in the observed trace trie (i.e., no observed abstract trace prefix realizes
the corresponding state-action-state steps), \TriCEGAR treats the witness as
\textit{unsupported-by-log} with respect to the currently observed behavior
and targets refinement at the earliest divergence point. This notion
of spuriousness is evidence-based: it is used to prioritize refinement towards
regions where the abstraction merges behaviorally distinct concrete contexts,
rather than to provide a completeness guarantee over unobserved executions.
Refinement splits the responsible leaf using a predicate that reduces
behavioral uncertainty, measured by IG on next actions.

In probabilistic settings, counterexamples can be represented as sets of
paths or critical subsystems rather than a single path. The planned
probabilistic refinement extracts the subset of abstract states and
transitions that contribute most to the violating probability mass, then
refines those states to tighten bounds and reduce spurious probability flow~\cite{han2009counterexample}.
Given the complexity and subtlety involved in applying probabilistic refinements, 
this work concentrates exclusively on the modeling aspects of the problem, 
while the refinement itself is deferred to future research. A complete visualization of the pipeline is presented in \autoref{fig:tricegar_pipeline}, accompanied by a detailed explanation of the approach in the appendices~\ref{app:tri} and \ref{app:three}.

\subsection{Implementation and Method}
\label{sec:implementation}

This subsection describes the prototype implementation, which supports the pipeline:
\emph{trace capture} $\rightarrow$
\emph{decision-tree abstraction} $\rightarrow$
\emph{MDP induction} $\rightarrow$
\emph{probabilistic model checking} $\rightarrow$
\emph{anomaly detection}.

\subsubsection{Lifecycle Interception and State Capture}
The implementation integrates as a monitoring feature that hooks into the
agent lifecycle 
using a generic \emph{event interceptor} on agent tool
invocations. At each tool invocation, the interceptor records 
(i) the action symbol (tool type), (ii) tool arguments and observations,
and (iii) a snapshot of selected internal variables. We formalize each
snapshot as
$
  s_c =
  \langle
    V_{\mathrm{goal}}, V_{\mathrm{check}}, V_{\mathrm{state}}
  \rangle.
$
\textbf{Goal variables} capture immutable task context (e.g., an issue
identifier or user request). \textbf{Checkpoint variables} capture low-dimensional
workflow flags (e.g., \texttt{testsPassed}).
\textbf{State variables} capture execution statistics and artifacts (e.g.,
step counters, diff sizes, exception counts, and summarized observations).
This partition supports two needs: checkpoint variables can label success and
failure states for model checking, while state variables provide features for
learning abstraction predicates. In the current prototype, the captured
variables are selected by developer configuration. The next step is supporting
annotation-driven selection to reduce configuration overhead.

\subsubsection{Trace Normalization and Storage}
Agent traces include high-dimensional and heterogeneous data. The
implementation stores captured variables as typed records: numeric values 
as floats or integers, booleans as booleans, collections with derived features  (cardinality, emptiness), and text as strings with optional embeddings.
For online processing, the continuous stream of events is segmented into discrete concrete transitions $s_c \xrightarrow{a} s'_c$, which are sequentially ingested into the trace trie and the (append only) trace log. This trie stores sequences of abstract states and actions, and maintains references to the underlying concrete records to support concretization checks and leaf-level refinement datasets.

\subsubsection{Decision-Tree Abstraction Construction}
The current prototype learns an initial predicate tree $T_0$ from the
recorded trace dataset. The training target is the next tool invocation, which
is available without manual labeling. Candidate predicates are generated
according to variable types:
(1) numeric thresholds from observed values, 
(2) boolean equality checks,
and (3) collection emptiness and cardinality thresholds.
The tree is built greedily using IG (in a process similar to ID3~\cite{quinlan1986induction}). To reduce overfitting in
small logs a configurable minimum IG threshold is used. These
limits are configurable parameters because they trade off model
size and sensitivity. In the current prototype, 
we rebuild the model periodically to include data from newer traces.

\subsubsection{Online MDP Construction and Update}
Let $S$ be the leaf set of $T$ and $A$ be the action alphabet (tools). The MDP
transition model maintains counts
$C(s,a,s')$ and $C(s,a)$.
Then, the transition probabilities are estimated by:
$
  P(s,a,s') = {C(s,a,s')}/{C(s,a)}.
$
In online settings, early estimates may be unstable due to sparsity.
We therefore use unsmoothed empirical frequencies and treat unobserved
transitions as absent, ensuring the model reflects only observed behavior,
which is suitable for anomaly detection and spuriousness checks.
Success and failure are defined via predicates over goal variables.
For instance, $\mathsf{success}$ may require
$\mathsf{testsPassed}=\mathsf{true} \wedge \mathsf{committed}=\mathsf{true}$,
while $\mathsf{failure}$ may correspond to a budget violation or terminal
error. These predicates support reachability specifications without
introducing an explicit reward model.

\subsubsection{Probabilistic Model Checking Interface}
The implementation exports the constructed MDP to the Storm model checker~\cite{dehnert_storm_2017} and evaluates
property templates. Our prototype implementation supports reachability-bound templates:
$
  P_{\max}=?\ [\ \Diamond\ \mathsf{success}\ ],
$
and
$
  P_{\min}=?\ [\ \Diamond\ \mathsf{failure}\ ]
$.
These quantities are used in two ways. First, they provide an interpretable
summary of expected behavior under uncertainty. Second, they can drive runtime
policies, such as warning when $P_{\min}(\Diamond\ \mathsf{failure})$ exceeds
a threshold.

\subsection{Anomaly Detection via Run Likelihood}
\label{sec:implementation:anomaly}

Model-based anomaly detection over probabilistic transition systems
commonly learns a transition model from historical executions and 
flags low-probability event sequences as anomalous.
We follow a similar methodology to previous work in this field~\cite{ye2004robustness,haque2017markov}. Using these principles, we are able to do both runtime and offline (post-run) anomaly detection, 
as detailed in~\autoref{app:anomaly}.

\paragraph{\textbf{Current status.}}
The current implementation includes the pipeline from lifecycle logging to $T_0$ learning, MDP induction, PCTL
reachability queries, and run likelihood computation. The
probabilistic CEGAR refinement loop and dynamic translation of high-level
properties to the induced model labels are ongoing work.


\section{Motivating Example}
\label{sec:example}

This section illustrates our framework's current workflow through a simple, representative agent that performs file read/write operations. The example demonstrates the complete workflow from state capture through anomaly detection and verification.

\textbf{State capture}
Consider an AI agent that performs file I/O operations, using the Koog Framework~\cite{jetbrainskoog_2025}. The agent maintains internal state variables that track its execution. AgentGuard intercepts each tool call, such as \lstinline|readFile()| and \lstinline|writeFile()|, using dynamic proxies, capturing the agent's state before and after each call. 

\textbf{Use Case}
The developer uses the framework to ensure the agent correctly handles file operations without entering infinite loops or deviating from expected workflows. For this, they annotate relevant state variables (such as \lstinline|filesWrittenCount|, \lstinline|lastFileRead|, \lstinline|iteration|) and tools, then run the agent to collect execution traces.

\textbf{Abstraction} 
Raw traces are high-dimensional, therefore \TriCEGAR automatically learns a predicate tree to partition these concrete snapshots into abstract states. For instance, the tree may learn to split states based on \lstinline|filesWritten > 0|. 

\textbf{Verification \& Anomaly Detection} 
Our implementation 
constructs an MDP from these abstract traces. Then, it model checks the PCTL properties $P_{max}(\text{success})$, $P_{min}(\text{fail})$, using STORM~\cite{dehnert_storm_2017}.
Simultaneously, it calculates the likelihood of the running execution trace against the historical model. A sequence with a low likelihood triggers an anomaly alert, enabling the monitor to identify issues such as 
unexpected loops or tool call patterns in real time.

\textbf{Refinement}
We generated 1000 baseline traces through randomized file operations (varying read/write ratios and trace lengths) and 1000 test traces with injected anomalies (e.g., extremely long/short traces, unbalanced read/write patterns, malformed paths). Our framework successfully detected both length-based and ratio-based anomalies. However, it also reported a high number of false positives for ratio-based properties. This suggests that some abstract states over-generalize the execution patterns. The counterexample-guided refinement step addresses this issue by identifying the abstract states 
that can be refined to eliminate spurious signals.


\section{Future Work, Discussion, and Limitations}
\label{sec:future}

\textbf{Probabilistic Counterexample-Guided Refinement}
The primary ongoing effort is implementing probabilistic CEGAR refinement.
In the probabilistic setting, refinement should target the portions of the
model that contribute to loose bounds or spurious witnesses.
This requires extracting counterexamples from model
checking and splitting abstraction leaves using predicates that separate
divergent next-action distributions at the divergence point.

\textbf{Dynamic Property Verification}
A second ongoing effort is translating high-level properties to induced
models created dynamically.
One direction is property templates that bind to labeled variables captured
at runtime (e.g., ``tests passed''), together with a mapping from tool types
to action symbols.
A second direction is learning label predicates from traces when labels are
not observable.

\textbf{Limitations}
The current approach is driven by observed traces.
If traces under-sample relevant behaviors, the induced MDP may not cover
unobserved outcomes.
Decision-tree abstractions can overfit when refinement is unconstrained.
These constraints motivate refinement strategies, resource bounds
(max depth, max leaves), and validation against held-out traces.


\section{Conclusion}
This paper proposes \TriCEGAR as a progression from runtime verification
frameworks that rely on manual state modeling.
\TriCEGAR learns abstractions from agent traces, constructs an MDP, and enables
probabilistic model checking queries that provide quantitative assurances.
We describe an implementation that supports trace capture, decision-tree
abstraction, MDP construction, reachability checking, and likelihood-based
anomaly detection.
Future work targets probabilistic counterexample-guided refinement and
property translation for dynamically induced models.

\newpage

\bibliographystyle{ACM-Reference-Format}
\bibliography{main}

\newpage
\appendix

\section{\TriCEGAR}
\label{app:tri}
\subsection{Overview}

We propose a framework for runtime verification / model learning in which the abstraction used for verification is constructed and refined automatically based on observed execution traces. The abstraction is realized as a \emph{predicate tree} (a decision tree over concrete-state variables) whose leaves define abstract states. An abstract model (e.g.\ an MDP) over these abstract states is learned from the traces and then verified against a property \(\phi\). If verification fails and the counterexample cannot be concretized inside the observed trace set, the abstraction is refined via IG–based splitting of the predicate tree. A prefix trie over concrete traces is used to support efficient concrete-to-abstract mappings and to detect spurious counterexamples.  

\subsection{Formal Definitions}

\paragraph{Concrete states and traces}  
Let \(\Sigma\) denote the set of all possible concrete states. Each concrete state is of the form  
\[
  s_c = \langle V_{goal}, V_{check}, V_{state} \rangle,
\]  
where \(V_{goal}, V_{check}, V_{state}\) are sets of variables whose types may be numeric, boolean, string, array/collection, etc.  

We observe a finite log  
\[
  L = \{ \sigma_i \mid i = 1, \dots, N \},
\]  
where each trace \(\sigma_i\) is a finite sequence of concrete states connected by tool (or action) invocations, i.e.  
\[
  \sigma_i = (s_{c,0} \xrightarrow{a_0} s_{c,1} \xrightarrow{a_1} \cdots \xrightarrow{a_{n-1}} s_{c,n} ).
\]  

\paragraph{Predicate tree (decision tree / tree automaton) abstraction}  
Let \(V\) denote the finite set of all variables appearing in concrete states. We define a \emph{predicate tree}  
\[
  T = (N, \mathit{root}, \mathit{lab}, \mathit{ch}_0, \mathit{ch}_1),
\]  
where:

\begin{itemize}
  \item \(N\) is the set of nodes, with a distinguished root node \(\mathit{root} \in N\).  
  \item For each internal node \(n \in N\), \(\mathit{lab}(n)\) is a predicate \(\pi_v\) over some variable \(v \in V\).  
  \item Each internal node \(n\) has exactly two children: \(\mathit{ch}_0(n)\) (the false branch) and \(\mathit{ch}_1(n)\) (the true branch).  
  \item Leaves \(\ell \in \mathrm{Leaves}(T)\) correspond to abstract states; define \(S_{abs} = \mathrm{Leaves}(T)\).  
\end{itemize}

Define the abstraction function \(\alpha_T: \Sigma \to S_{abs}\) as follows: given \(s_c \in \Sigma\), traverse \(T\) from the root; for each internal node \(n\) with predicate \(\pi_v\), evaluate \(\pi_v(s_c)\), and follow \(\mathit{ch}_b(n)\) where \(b =\) false or true accordingly. The leaf reached is \(\alpha_T(s_c)\).  

Thus \(T\) can be interpreted both as a classical decision tree over variables in \(V\), and as a (deterministic) tree automaton whose states are the nodes of \(T\), whose transitions correspond to predicate evaluations (false = 0 / true = 1), and whose accepting states are the leaves. Abstract states are identified by root-to-leaf paths (bit-strings of predicate outcomes).

\paragraph{Concrete-trace trie}  
We build a prefix trie \(\mathcal T_{\mathrm{trie}}\) over the observed (abstract) traces \(L\). Each trace is encoded as a word over an alphabet capturing tool invocations. The trie \(\mathcal T_{\mathrm{trie}} = (W, \varepsilon, \mathit{succ})\) satisfies:

\begin{itemize}
  \item \(W\) is the set of all prefixes of all trace-words in \(L\).  
  \item \(\varepsilon\) is the empty prefix (root).  
  \item If \(w \in W\) is a prefix and \(a\) is an action, then \(\mathit{succ}(w, a) = w \cdot a\) if there exists a trace in \(L\) whose prefix is \(w\) followed by action \(a\).  
\end{itemize}

This trie enables efficient enumeration of concrete traces (or concrete-state sequences) and efficient mapping from concrete states (or trace prefixes) to their abstract states under \(\alpha_T\).  

\paragraph{Abstract model (graph/MDP)}  
Given abstraction \(\alpha_T\) and trace log \(L\), we define an abstract model  
\[
  \mathcal M = (S_{abs}, A, P, R),
\]  
where:

\begin{itemize}
  \item \(S_{abs} = \mathrm{Leaves}(T)\).  
  \item \(A\) is the set of actions / tool invocations observed in \(L\).  
  \item For each \(s, s' \in S_{abs}\) and action \(a \in A\), define transition probability  
    \[
      P(s, a, s') = \frac{ \bigl| \{ (s_c \xrightarrow{a} s'_c) \in L \mid \alpha_T(s_c)=s,\; \alpha_T(s'_c)=s' \} \bigr| }{ \bigl| \{ s_c \mid \alpha_T(s_c)=s \text{ and next action } = a \} \bigr| }.
    \]  
  \item \(R\) (optional): a reward or labeling function, aggregated from concrete transitions if needed. (I include this exclusively for completeness)
\end{itemize}

\subsection{Initialization: Building the Initial Predicate Tree from Data}

Rather than starting from a trivial abstraction, we construct the initial predicate tree \(T_0\) by applying a decision-tree construction procedure on the set of concrete states observed in \(L\), using an IG criterion (entropy-based) to guide splits, and using variables \(V\) as features.  

\medskip
\noindent\textbf{Procedure \textsc{BuildInitialTree}:}  

\begin{enumerate}
  \item Let  
    \[
      B = \{ s_c \in \Sigma \mid s_c \text{ appears in some trace in } L \}
    \]  
    be the multiset (or set) of concrete states observed.  
  \item Define a classification function \(\mathit{class}: B \to \mathcal C\), where \(\mathcal C\) is a set of behaviorally relevant classes; for example, classify each state by the next action in its trace (tool invocation), or by the trace-suffix signature, or by whether continuing execution eventually leads to violation of property \(\phi\).  
  \item Recursively build a decision tree over \(V\), as follows:  

    \begin{itemize}
      \item For current subset \(B' \subseteq B\), compute the entropy  
        \[
          \Eta(B') = - \sum_{c \in \mathcal C} p_c \log_2 p_c,
        \]  
        where \(p_c = \bigl|\{ s \in B' \mid \mathit{class}(s) = c \}\bigr| / |B'|\).  
      \item For each candidate variable \(v \in V\), for each plausible predicate \(\pi_v\) over \(v\) (depending on type: threshold for numeric, non-emptiness for collections, boolean, string-based predicate, etc.), partition \(B'\) into  
        \[
          B'_0 = \{ s \mid \pi_v(s) = \mathrm{false} \}, \quad
          B'_1 = \{ s \mid \pi_v(s) = \mathrm{true} \}.
        \]  
        Compute  
        \[
          \mathrm{IG}(B',\pi_v) = \Eta(B') - \left( \frac{|B'_0|}{|B'|} \Eta(B'_0) + \frac{|B'_1|}{|B'|} \Eta(B'_1) \right).
        \]  
      \item Select \(\pi_{v^*}\) that maximizes \(\mathrm{IG}\). If \(\max \mathrm{IG} \le \gamma\) (some threshold), stop and make current node a leaf. Otherwise, create internal node with \(\pi_{v^*}\) and recursively build subtrees on \(B'_0, B'_1\), removing \(v^*\) from future splits along those branches to avoid re-splitting on same variable.  
    \end{itemize}

  \item The result is a predicate tree \(T_0\) whose leaves partition \(B\); that defines abstraction \(\alpha_{T_0}\).  
  \item Build the trace trie \(\mathcal T_{\mathrm{trie}}\) over \(L\).  
  \item Build the initial abstract model \(\mathcal M_0\) via \(\alpha_{T_0}\).  
\end{enumerate}

After initialization, we obtain a non-trivial abstraction (not just a single abstract state) whose structure already reflects distinctions in behavior observed in the data.  

\subsection{Verification and Refinement Loop}

We embed this initialization into a modified CEGAR-style loop:

\begin{enumerate}
  \item From the current abstraction \(\alpha_T\) and log \(L\), build abstract model \(\mathcal M\).  
  \item Verify property \(\phi\) on \(\mathcal M\).  
    \begin{itemize}
      \item If \(\mathcal M \models \phi\), conclude “verified (under abstraction)”.  
      \item Otherwise obtain abstract counterexample \(\tau_{ce}\).  
    \end{itemize}
  \item Use the concrete-trace trie \(\mathcal T_{\mathrm{trie}}\) to attempt to concretize \(\tau_{ce}\).  
    \begin{itemize}
      \item If a matching concrete trace exists (or plausible extension), report a real counterexample.  
      \item Otherwise the counterexample is spurious.  
    \end{itemize}
  \item For each abstract state \(s_{abs}\) along \(\tau_{ce}\) that contributes to the spuriousness, collect the set  
    \[
      B = \{ s_c \in \Sigma \mid \alpha_T(s_c) = s_{abs}, s_c \in L \}.
    \]  
    Use the same IG splitting procedure over variables in \(V\) (excluding those already used along the path) to choose a new predicate \(\pi_v\) that maximally reduces behavioral uncertainty. Replace the leaf node corresponding to \(s_{abs}\) by an internal node with \(\pi_v\), yielding refined tree \(T'\), update \(\alpha_T := \alpha_{T'}\).  
  \item Re-abstract \(L\), rebuild \(\mathcal M\), and repeat.  
  \item Terminate when either \(\phi\) holds on \(\mathcal M\), a real counterexample is found, or no beneficial split can be found (IG below threshold), or a resource bound (max depth, max leaves, max iterations) is reached.  
\end{enumerate}

\subsection{Remarks on Soundness and Limitations}

This framework adopts the general philosophy of Counterexample-Guided Abstraction Refinement (CEGAR), where abstraction simplification and iterative refinement are used to balance tractability and precision. The key difference is that the abstraction (predicate tree) is not supplied manually, but learned from data via a decision-tree algorithm at initialization and refined automatically.  

Because abstraction groups together concrete states into abstract states, the abstract model over-approximates the set of behaviors observed in \(L\). Thus if \(\mathcal M \models \phi\), then all observed traces satisfy \(\phi\). However, the abstraction may miss unobserved concrete behaviors; completeness with respect to all possible concrete executions is not guaranteed.  

Moreover, the quality of the abstraction depends on the representativeness of \(L\), the choice of classification function for entropy, and the choice of splitting predicates. Poor choices may lead to overfitting (over-specialized abstraction) or underfitting (spurious counterexamples persist).  

Nevertheless, this method offers a systematic, data-driven way to build and refine abstractions automatically, reducing manual predicate engineering.

\section{Three-Dimensional Abstraction Structure}
\label{app:three}

The previous sections define three separate data structures: the predicate tree \(T\) that implements the abstraction function \(\alpha_T\), the abstract model graph \(\mathcal M\), and the concrete-trace trie \(\mathcal T_{\mathrm{trie}}\). In an implementation, these structures should not be treated as independent objects. Instead, we define a single three-dimensional abstraction structure that maintains explicit links between:

\begin{itemize}
  \item tree nodes in \(T\) (data abstraction dimension),
  \item graph vertices in \(\mathcal M\) (dynamics dimension),
  \item trie nodes in \(\mathcal T_{\mathrm{trie}}\) (trace dimension).
\end{itemize}

This section specifies this combined structure and the main operations on it.

\subsection{Combined Abstraction Structure}

We assume that each of the three underlying structures uses stable identifiers:

\begin{itemize}
  \item \(\mathit{TreeId}\): identifiers for nodes in the predicate tree \(T\),
  \item \(\mathit{GraphId}\): identifiers for abstract states (vertices) in \(\mathcal M\),
  \item \(\mathit{TrieId}\): identifiers for nodes in the concrete-trace trie \(\mathcal T_{\mathrm{trie}}\).
\end{itemize}

\begin{definition}[Abstract State Handle]
An \emph{abstract state handle} is a record
\[
  h = \langle \mathit{tree}, \mathit{graph}, \mathit{endpoints} \rangle
\]
where:
\begin{itemize}
  \item \(\mathit{tree} \in \mathit{TreeId}\) is the identifier of a leaf node in \(T\),
  \item \(\mathit{graph} \in \mathit{GraphId}\) is the identifier of the corresponding vertex in \(\mathcal M\),
  \item \(\mathit{endpoints} \subseteq \mathit{TrieId}\) is the set of trie nodes that correspond to trace prefixes whose last concrete state currently abstracts to this leaf.
\end{itemize}
\end{definition}

The endpoints set is implementation level bookkeeping. For each abstract state we keep a compact representation of ``where in the trie'' the concrete states mapped to it end up.

\begin{definition}[Three-dimensional abstraction structure]
The three-dimensional abstraction structure is a tuple
\[
  \mathcal D = \langle T,\ \mathcal M,\ \mathcal T_{\mathrm{trie}},\ H,\ \mathit{mapTree},\ \mathit{mapGraph},\ \mathit{mapTrie} \rangle
\]
where:
\begin{itemize}
  \item \(T\) is the predicate tree defined earlier,
  \item \(\mathcal M = (S_{abs}, A, P, R)\) is the abstract model graph,
  \item \(\mathcal T_{\mathrm{trie}}\) is the concrete-trace trie,
  \item \(H\) is a finite set of abstract state handles,
  \item \(\mathit{mapTree}: \mathit{TreeId} \rightharpoonup H\) is a partial function mapping each leaf-identifier of \(T\) to its handle,
  \item \(\mathit{mapGraph}: \mathit{GraphId} \rightharpoonup H\) is a partial function mapping each abstract-state vertex in \(\mathcal M\) to its handle,
  \item \(\mathit{mapTrie}: \mathit{TrieId} \rightharpoonup H\) is a partial function assigning to each trie node that ends at a concrete state the handle of the abstract state to which that concrete state currently maps.
\end{itemize}
\end{definition}

Intuitively, \(T\) and \(\mathcal M\) define the abstraction and transitions, \(\mathcal T_{\mathrm{trie}}\) stores concrete traces, and \(H\) plus the three maps maintain consistency between the three dimensions.

\subsection{Representation Invariants}

The following invariants should hold for every valid instance of \(\mathcal D\):

\begin{description}
  \item[I1 (Leaf–graph consistency).] For every handle \(h \in H\),
  \[
    h.\mathit{tree} \in \mathrm{Leaves}(T)
    \quad\text{and}\quad
    h.\mathit{graph} \in S_{abs},
  \]
  and \(\mathit{mapTree}(h.\mathit{tree}) = h\), \(\mathit{mapGraph}(h.\mathit{graph}) = h\).

  \item[I2 (Trie endpoint consistency).] For every handle \(h \in H\) and every \(u \in h.\mathit{endpoints}\),
  \[
    \mathit{mapTrie}(u) = h,
  \]
  and the concrete state at trie node \(u\) is abstracted by \(\alpha_T\) to the leaf node with identifier \(h.\mathit{tree}\).

  \item[I3 (Coverage).] Every abstract state in \(S_{abs}\) has exactly one handle, and every leaf in \(T\) has exactly one handle. Formally:
  \[
    \forall s \in S_{abs}.\ \exists! h \in H.\ h.\mathit{graph} = s,
  \]
  and
  \[
    \forall \ell \in \mathrm{Leaves}(T).\ \exists! h \in H.\ h.\mathit{tree} = \ell.
  \]

  \item[I4 (Graph state set).] The set of vertices in \(\mathcal M\) coincides with the set of graph identifiers used by handles:
  \[
    S_{abs} = \{ h.\mathit{graph} \mid h \in H \}.
  \]
\end{description}

\section{Run-Likelihood Anomaly Detection Details}
\label{app:anomaly}

\paragraph{Run likelihood.}
Let the induced model be an MDP with transition probabilities
$P(s,a,s')$ over abstract states and tool actions.
For an observed abstract run
$\rho = (s_0,a_0,s_1,\ldots,a_{n-1},s_n)$,
we define its (model) likelihood as
\begin{equation}
  \Pr(\rho) \;=\; \prod_{i=0}^{n-1} P(s_i,a_i,s_{i+1}).
  \label{eq:run_likelihood}
\end{equation}
In practice we operate on log-likelihoods for numerical stability:
\begin{equation}
  \ell(\rho) \;=\; \log \Pr(\rho)
  \;=\; \sum_{i=0}^{n-1} \log P(s_i,a_i,s_{i+1}).
  \label{eq:run_loglik}
\end{equation}

\paragraph{Offline (run-level) anomaly detection.}
Given a history of completed runs $\mathcal{R}$, we can compute the sample
mean and variance of the run log-likelihoods:
\begin{equation}
  \mu \;=\; \frac{1}{|\mathcal{R}|}\sum_{\rho \in \mathcal{R}} \ell(\rho),
  \qquad
  \sigma^2 \;=\; \frac{1}{|\mathcal{R}|-1}\sum_{\rho \in \mathcal{R}}
  \bigl(\ell(\rho)-\mu\bigr)^2.
  \label{eq:run_stats}
\end{equation}
A simple detector flags a run as anomalous if $\ell(\rho)$ falls outside an
acceptance region. Under a normal approximation for $\ell(\rho)$, a
two-sided $(1-\alpha)$ interval is
$[\mu - z_{1-\alpha/2}\sigma,\ \mu + z_{1-\alpha/2}\sigma]$.
Since anomalies typically correspond to \emph{unexpectedly low} likelihood,
we use a one-sided rule:
\begin{equation}
  \rho \text{ is anomalous if }
  \ell(\rho) < \mu - z_{1-\alpha}\sigma.
  \label{eq:offline_threshold}
\end{equation}
When the normal approximation is not appropriate, the same rule can be
implemented using empirical quantiles of $\{\ell(\rho)\}_{\rho\in\mathcal{R}}$.

\paragraph{Online (runtime) anomaly detection via prefix-conditioned statistics.}
A direct application of \eqref{eq:offline_threshold} at runtime is not
meaningful because $\ell(\rho)$ decreases in magnitude as the run length
increases (it is a sum over more transitions). We therefore maintain
\emph{prefix-conditioned} statistics at a set of checkpoints
$\mathcal{K} = \{k_1,k_2,\ldots\}$ (e.g., $10,20,30,\ldots$ jumps).

For any run $\rho$ with length $n$, define its length-$k$ prefix
$\rho_{\le k}$ for $k \le n$ and the corresponding prefix log-likelihood:
\begin{equation}
  \ell_k(\rho) \;=\; \ell(\rho_{\le k})
  \;=\; \sum_{i=0}^{k-1} \log P(s_i,a_i,s_{i+1}).
  \label{eq:prefix_loglik}
\end{equation}
For each checkpoint $k \in \mathcal{K}$, we compute statistics using only
historical runs that are at least $k$ steps long:
\begin{equation}
  \begin{aligned}
  \mathcal{R}_{\ge k} \;=\; \{\rho \in \mathcal{R} \mid |\rho| \ge k\},
  \\
  \mu_k \;=\; \frac{1}{|\mathcal{R}_{\ge k}|}\sum_{\rho \in \mathcal{R}_{\ge k}}
  \ell_k(\rho),
  \\
  \sigma_k^2 \;=\; \mathrm{Var}\bigl(\{\ell_k(\rho)\}_{\rho \in \mathcal{R}_{\ge k}}\bigr).
  \label{eq:prefix_stats}
  \end{aligned}
\end{equation}
This implements the following normalization for variable-length traces:
runs shorter than $k$ are excluded, and runs longer than $k$ are truncated
to their first $k$ transitions before computing $\ell_k(\rho)$.

During an ongoing execution, when the current run reaches a checkpoint
$k \in \mathcal{K}$ we compute $\ell_k(\rho_{\mathrm{cur}})$ and issue a warning if
it is outside the acceptance region for that checkpoint. A one-sided
low-likelihood warning rule is:
\begin{equation}
  \text{warn at step } k \text{ if }
  \ell_k(\rho_{\mathrm{cur}}) < \mu_k - z_{1-\alpha}\sigma_k.
  \label{eq:online_threshold}
\end{equation}
This provides a runtime anomaly detector whose threshold is comparable across
execution time because it conditions on the same prefix length.

\end{document}